\documentclass[conference]{IEEEtran}
\IEEEoverridecommandlockouts
\usepackage{cite}
\usepackage{amsmath,amssymb,amsfonts}
\usepackage{algorithmic}
\usepackage{graphicx}
\usepackage{textcomp}
\usepackage{subcaption}
\usepackage{hyperref}
\usepackage{xcolor}
\usepackage{soul}
\def\BibTeX{{\rm B\kern-.05em{\sc i\kern-.025em b}\kern-.08em
    T\kern-.1667em\lower.7ex\hbox{E}\kern-.125emX}}
\begin{document}

\title{Robots with Different Embodiments Can Express and Influence Carefulness in Object Manipulation
\thanks{Alessandra Sciutti is supported by a Starting Grant from the European Research Council (ERC) under the European Union’s Horizon 2020 research and innovation programme. G.A. No 804388, wHiSPER. This work is also supported by the European Commission within the Horizon 2020 Framework (CHIST-ERA, 2014-2020), project InDex,
and by the grant n. FA8655-20-1-7035 founded by AFOSR. 
This work has been partially carried out at the Machine Learning Genoa (MaLGa) center, Universita` di Genova (IT).
}
}

\author{
\IEEEauthorblockN{Linda Lastrico\IEEEauthorrefmark{2}\IEEEauthorrefmark{4}, Luca Garello\IEEEauthorrefmark{2}\IEEEauthorrefmark{4}, Francesco Rea\IEEEauthorrefmark{4}, Nicoletta Noceti\IEEEauthorrefmark{2},\\Fulvio Mastrogiovanni\IEEEauthorrefmark{2}, Alessandra Sciutti\IEEEauthorrefmark{3}, Alessandro Carf\`i\IEEEauthorrefmark{2}}

\IEEEauthorblockA{\IEEEauthorblockA{\IEEEauthorrefmark{2}Department of Informatics, Bioengineering, Robotics, and Systems Engineering (DIBRIS),\\ University of Genoa, Italy
\IEEEauthorblockA{\IEEEauthorrefmark{4}Robotics, Brain and Cognitive Science Department (RBCS),\\ Italian Institute of Technology, Genoa, Italy\\
\IEEEauthorrefmark{3}Cognitive Architecture for Collaborative Technologies Unit (CONTACT),\\ Italian Institute of Technology, Genoa, Italy}
}
}
}

\maketitle

\begin{abstract}
Humans have an extraordinary ability to communicate and read the properties of objects by simply watching them being carried by someone else. This level of communicative skills and interpretation, available to humans, is essential for collaborative robots if they are to interact naturally and effectively. For example, suppose a robot is handing over a fragile object. In that case, the human who receives it should be informed of its fragility in advance, through an immediate and implicit message, i.e., by the direct modulation of the robot's action. This work investigates the perception of object manipulations performed with a communicative intent by two robots with different embodiments (an iCub humanoid robot and a Baxter robot). We designed the robots' movements to communicate carefulness or not during the transportation of objects. We found that not only this feature is correctly perceived by human observers, but it can elicit as well a form of motor adaptation in subsequent human object manipulations. In addition, we get an insight into which motion features may induce to manipulate an object more or less carefully.

\end{abstract}


\section{Introduction}
To establish an effective interaction between humans and robots, the robot must generate readable and legible movements \cite{legibility}. As a  broad review by Venture and Kuli\'c points out \cite{kulicReview}, robots can exploit their embodiment to be communicative while performing a task to transmit more information to their partner. In particular, research on implicit communication focuses mainly on two goals: conveying the robot's intentionality or affective/emotional attitude. Another context where robots' communicative potential has been explored is the co-verbal one: generating human-like gestures correlated with speech makes the robot more understandable and perceived positively \cite{CommunicativeGestures, CommunicativeGestures2}. However, humans tend to exchange implicit information, often unconsciously, also in other contexts. For instance, when dealing with object transportation, it has been shown that the weight we are carrying naturally modulates our movements \cite{weightFlanagan,velWeight}. Humans, even children, can estimate the weight of an object being transported by someone else \cite{massPerception, sciutti:weightChildren}. Also, the fragility can influence the arm kinematics during the manipulation: individuals will actively change their motor plan to preserve the object's integrity \cite{carefulnessBillard}. The modulation of action kinematics, which is functional for the proper task completion, also becomes useful for the observers - implicitly revealing the properties of the manipulated objects. 
To provide the same amount of information about the object they are carrying, robots should modulate their kinematics accordingly, improving the task's safety and efficiency. For what regards the carefulness associated with a transported object, recent studies proposed solutions to generate human-inspired communicative transportation motions \cite{carefulnessBillard,specialIssue}. Although it has been proved that robots can replicate those motions preserving the original kinematic qualities, we have no evidence that this allows robots to communicate the carefulness associated with an object to a human. To the best of our knowledge, there are no works in literature where the perception of the robot arm kinematics modulation, associated with the transport of objects with specific characteristics, has been investigated. In this context, the research questions targeted by our study are the following:
\begin{itemize}
    \item[] \textit{1\textsuperscript{st} Research Question} \textit{(RQ1)}: Are human-derived motions enough for the robot to communicate the carefulness required to transport an object?
    \item[] \textit{2\textsuperscript{nd} Research Question} \textit{(RQ2)}: Can the carefulness demonstrated by a robot influence how humans handle objects?
    \item[] \textit{3\textsuperscript{rd} Research Question} \textit{(RQ3)}: Do different robot embodiments affect the carefulness perceived by humans?
\end{itemize}
To explore these research questions, we deployed an already proposed solution \cite{ICDL2021_GAN, specialIssue} to generate transportation motions on two different robotic platforms: iCub (a humanoid) by the Italian Institute of Technology \cite{iCub} and Baxter (a two-arms manipulator) by Rethink Robotics \cite{baxter}. Due to their different conformation, these two platforms are suitable to explore the possible effect of the embodiment \textit{(RQ3)}. Videos recorded with the two robots have been used to investigate \textit{(RQ1)} through an online questionnaire and \textit{(RQ2)} with an in-presence experiment.

\section{Methods}\label{sec:methods}
We decided to address \textit{(RQ1)} by preparing a questionnaire including a few videos of Baxter and iCub transporting a blurred object, with what we programmed as a Careful (C) or Not Careful (NC) attitude. Participants were asked to evaluate the movement carefulness, on a five-points Likert scale from ``not careful" to ``careful" (carefulness score). Moreover, we presented a list of common-use objects and asked how careful we should be in transporting such items on the same scale. This is to clarify the concept of carefulness and to understand which attributes make an object more or less delicate to carry from the participant's perspective.\\
Concerning \textit{(RQ2)}, we opted for an in-presence experiment. In this case, participants transported a sensorized cube after watching videos of one of the robots carrying the same item, again either with a C or NC attitude. The experiment allowed us to investigate if participants, with no prior knowledge of the robots' attitude, would be influenced by their movement to the point of modulating their kinematics.\\
Both the experiments, relying on videos realized with different robotics platforms, will also allow quantifying \textit{(RQ3)}.
In the following part, we will give a short overview of how we generated iCub and Baxter movements. Then, we will describe the questionnaire. Finally, we will present the in-presence experiment setup.
\subsection{Generating communicative movements on robots}
\label{sec:movementGen}
\begin{figure}[h!]
 \centering
    \begin{subfigure}[]{1\columnwidth}
    \centering
    \includegraphics[width=1\columnwidth]{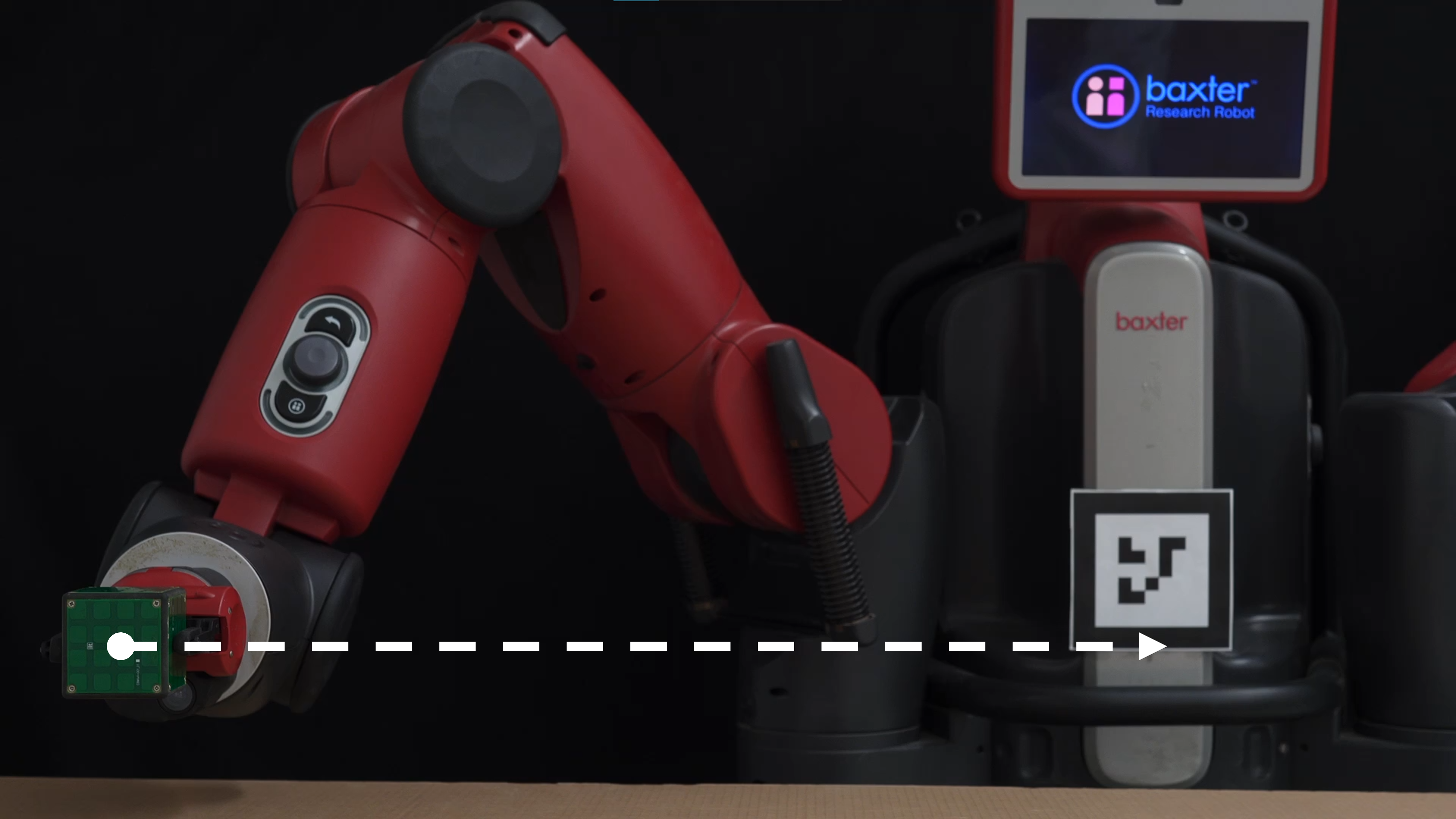}
    \caption{Baxter robot}
    \label{fig:baxter}
    \end{subfigure}
\\
\vspace{0.3cm}
\begin{subfigure}[]{1\columnwidth}
    \centering
    \includegraphics[width=1\columnwidth]{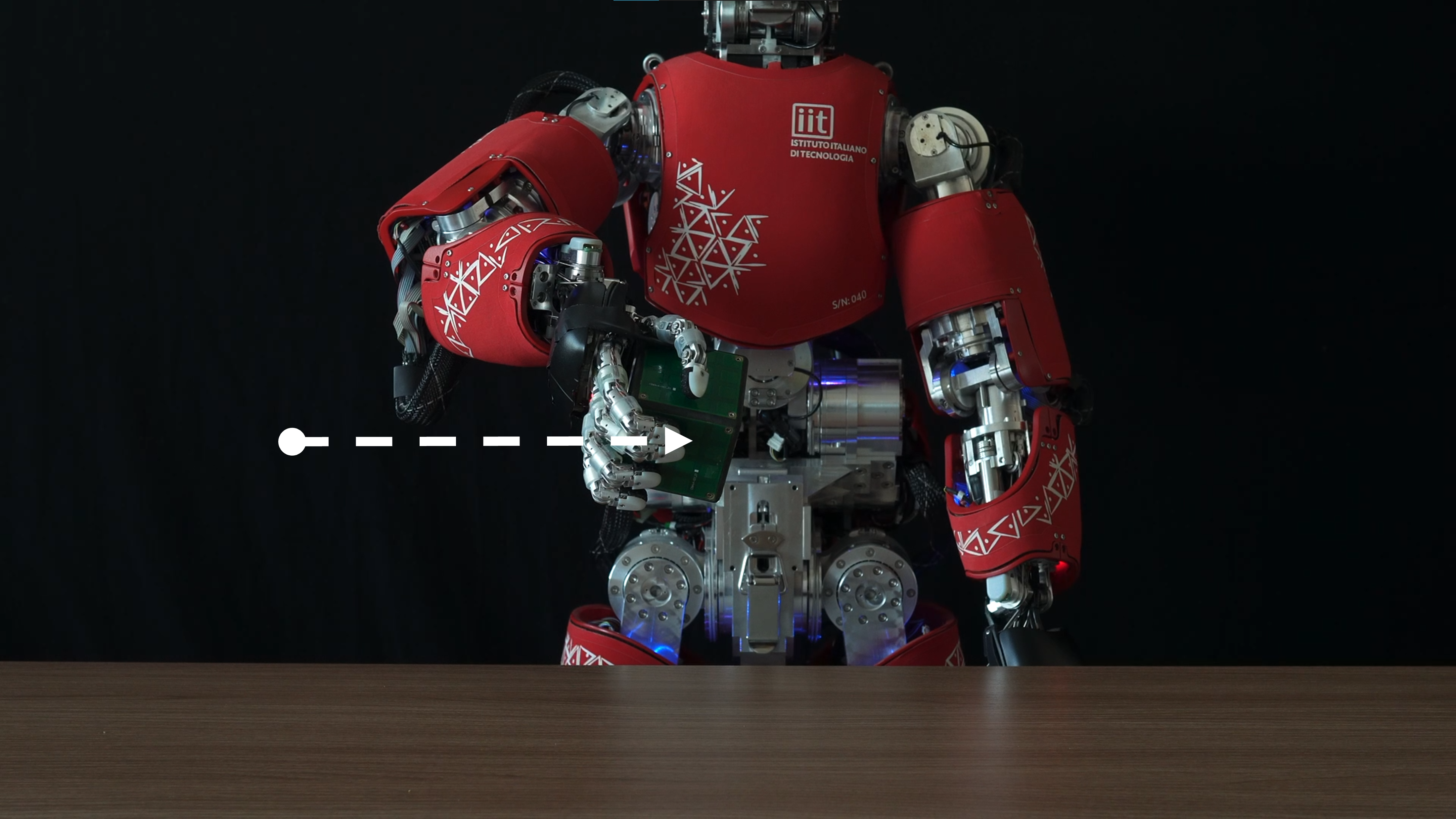}
    \caption{iCub robot}
    \label{fig:icub}
\end{subfigure}
\caption{Frames of the videos portraying the transport of a green cube along the frontal plane. Baxter (\ref{fig:baxter}) is at the beginning of the movement, while iCub (\ref{fig:baxter}) at its end. The dashed arrows represent the trajectory of the movement, which starts from the white dot and develops in the direction of the arrow}
\label{fig:robots}
\end{figure}
Previous literature showed that human movements naturally embed implicit information that allows an observer to deduce additional properties of the action being witnessed. For instance, information on the weight or the carefulness of the handled object can be deduced from the movement velocity \cite{weightFlanagan, velWeight, sciutti:weightChildren, icsrLinda, carefulnessBillard}. For this reason, in previous studies \cite{ICDL2021_GAN, specialIssue}, we focused on generating communicative robots' movement by modulating the velocity of their end-effector. Our goal was not to simply replicate human kinematics, but to exploit Generative Adversarial Networks to produce novel and meaningful velocity profiles associated with a careful, or not so careful, manipulation. We will now explain how, starting from one desired velocity profile, we controlled the robots to follow it.
\subsubsection{Baxter}
We used MoveIt!, a ROS motion planning framework widely used by several robotics systems\cite{moveit}, to control Baxter. We remapped the selected velocity profile to meet the intended trajectory length, keeping constant the temporal duration of the action while modulating the magnitude of the velocity. Due to the Moveit! controller implementation, we used a fixed spatial step to cover the trajectory while modulating the temporal execution. Hence, MoveIt! planned a trajectory defined by a sequence of poses in the joints space. We imposed the time with which each spatial point was supposed to be reached to determine the velocity norm of each spatial step, mapping the desired velocity profile. Baxter trajectories had a range of 60 cm.
\subsubsection{iCub}
We controlled the iCub robot with its default ``Cartesian Controller" \cite{pattacini}. However, differently from Baxter, on iCub, the end-effector controller required to fix the execution time of each sub-step, varying the Euclidean distance between them. Due to its kinematic configuration, iCub trajectories were shorter, around 20 cm, and, since the movement's original duration was preserved, slower than Baxter ones.\\
To prepare the stimuli for the questionnaire and the in-presence experiment, we recorded videos of iCub and Baxter moving the same object, a green cube, along the frontal plane. In particular, we used 2 distinct velocity profiles per condition (C/NC) per robot, resulting in 8 videos for the online questionnaire, while 6 videos for each combination for the experiment, resulting in 24 different videos. 
The movements in the videos where characterized by a maximum velocity reached by the end-effector in the Not Careful condition of $0.281 \pm  0.037$ $m/s$ for iCub, and of $0.871 \pm  0.080$ $m/s$ for Baxter. In the Careful videos instead, iCub reached a maximum velocity of $0.145 \pm  0.061$ $m/s$, while Baxter of $0.312 \pm  0.049 $ $m/s$. Regarding the end-effector mean velocity, we report in the Not Careful condition for iCub $0.137 \pm  0.023$ $m/s$, while for Baxter $0.348 \pm  0.129$ $m/s$. In the Careful videos, iCub had a mean velocity of $0.076 \pm  0.032$ $m/s$, while Baxter of $0.132 \pm  0.048 $ $m/s$.
As an example, in Figure \ref{fig:robots} are shown two images taken from the videos.
\begin{figure*}[h!]
 \centering
    \begin{subfigure}[]{.35\textwidth}
    \centering
    \includegraphics[width=1\linewidth]{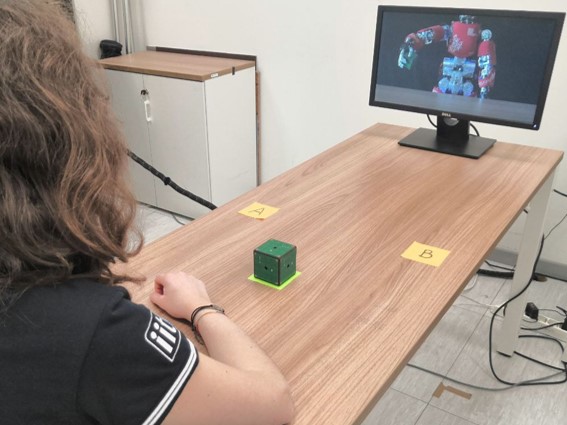}
    \caption{Participant watching one of the iCub's videos where it transports the sensorized cube}
    \label{fig:setupVideo}
    \end{subfigure}
\hspace{1em}
\begin{subfigure}[]{.35\textwidth}
    \centering
    \includegraphics[width=1\linewidth]{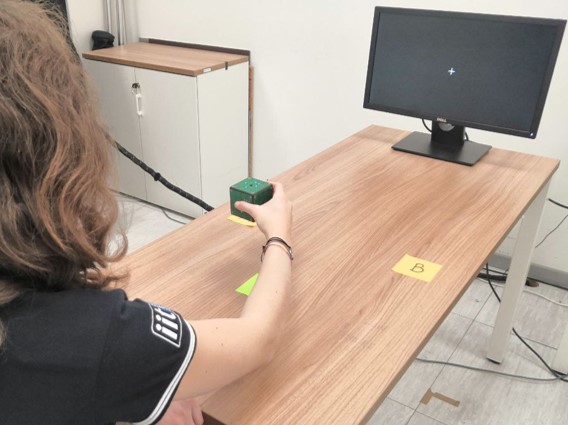}
    \caption{The participant places the cube in the indicated position, after watching the video}
    \label{fig:setupMov}
\end{subfigure}
\caption{Example of the experimental setup where the participants were asked to manipulate the cube (\ref{fig:setupMov}) after watching a video of one of the robots transporting it (\ref{fig:setupVideo})}
\label{fig:setup}
\hfill
\end{figure*}
\subsection{Online questionnaire}
At the beginning of the questionnaire, we asked participants their age, gender, and general knowledge about robotics, with possible answers going from 1 (``None - I have no idea what it is") to 5 (``Professional - I work, or worked, in the area of robotics"). 
In the first part, we asked them to observe the 8 C/NC videos of Baxter and iCub moving their end-effector, presented in random order. In the questionnaire, the videos of Baxter's and iCub's movements had the same setting as in Figure \ref{fig:robots}, but the end-effector was blurred. For each video, participants had to rate from 1, ``Not careful" to 5 ``Careful" the observed action, answering the question ``How attentive was the robot in moving the object?". Indeed, they were told the robot was moving a cubic container, and since the end-effector was blurred, they could not see what was inside it. 
In the second part, we proposed a list of 12 items and asked to imagine they would have to be carried around. The instructions to this part stated `For each of them, evaluate how much attention and care should be associated with the manipulation", again on a scale from 1, ``Not careful" to 5 ``Careful". We chose the items among everyday use objects, and they were: glass full of water till the brim, rubber ball, pair of scissors, worn out pelouche, lit candle, wooden cube, origami, plastic bottle, crystal cup, USB charger, face mask and a pair of glasses.\\
49 people (age: $25.5\pm5.7$, 25 females, 23 males, 1 non-binary/genderfluid) freely chose to answer the questionnaire.

\subsection{In-presence experimental setup}
We ran an in-person experiment to see if, beyond understanding the carefulness in the robots' movements when explicitly questioned about it, observing the robots' actions could influence how a person carries an object.
The experiment involved 11 healthy right-handed subjects that voluntarily agreed to participate in the data collection (8 females, 3 males, age: $26.5\pm 3.1$). All volunteers are members of our organization, but none is implicated in the research. \\
For this study, we used 6 videos per robot per condition, for a total of 24 trials, reproduced in casual order and with casual direction of the robot movement: always on the frontal plane, but going either from right to left or vice versa, by mirroring the video.\\
During the experiment, participants were asked to look carefully at the video displayed on the monitor in front of them (see Figure \ref{fig:setupVideo}) and afterward, following the instruction given by a synthetic voice, to move the cube they would find in front of them either to position A or B, marked by a colored paper square on the table (see Figure \ref{fig:setupMov}). The object they had to move was the iCube, a green sensorized cube of side 5 cm \cite{icube}.\\
Each participant was asked to read the same instructions, which were carefully designed so as not to introduce bias in the experiment. In particular, we explained that they would watch videos of the robots transporting the iCube. We used several prototypes of this cube, aesthetically indistinguishable but more or less delicate depending on their assembly state and internal components. At this point, we showed a box containing three iCubes to be used for the experiment. Even though we actually used only one of them, the purpose of having three was to avoid participants assuming a priori a binary interpretation of the videos (C or NC, even though this was exactly what we implemented in the robot actions). At the end of each trial, the participants were asked to close their eyes, for the experimenter to place a new cube in the starting position without influencing them with how the cube was moved. A sound informed the subject to open their eyes and prepare for watching a new video. We explained that the purpose of the experiment was to simulate a collaboration between the robot in the video and the participant. So, they were asked to imagine that, in every trial, the cube on the table was the one manipulated by the robot in the video, who virtually deposited it on the table for them to grab. It should be noted that to avoid a simple copy of the observed movement, while the robots' actions occurred in the frontal plane, the participants moved the iCube in the sagittal direction. Before starting with the video reproduction, we ran a baseline of 6 trials with no stimuli, for the participants to familiarize with moving the cube from the starting position to the ones in A or B.
To precisely record the hand kinematic during the transport of the cube, we used active infrared markers from the Optotrak Certus\textsuperscript{\textregistered}, NDI, motion capture (MoCap) system, which gave the 3D position of each marker with respect to a common reference frame, with a frequency of 100 Hz. We positioned four markers on the metacarpal bones of the right hand and considered for further analyses the most visible one in each trial.
For the purpose of this study and the sake of space, we report the results of the analysis of the kinematics data coming from the motion capture system and not those recorded through the iCube.
\section{Results}\label{sec:results}
To analyse the acquired data, we used Jamovi software \cite{jamovi}, in particular the GAMLj module \cite{jamovi_mixed} for mixed models. 
\subsection{Online questionnaire}
\begin{figure}[h!]
\centerline{\includegraphics[width=1\columnwidth]{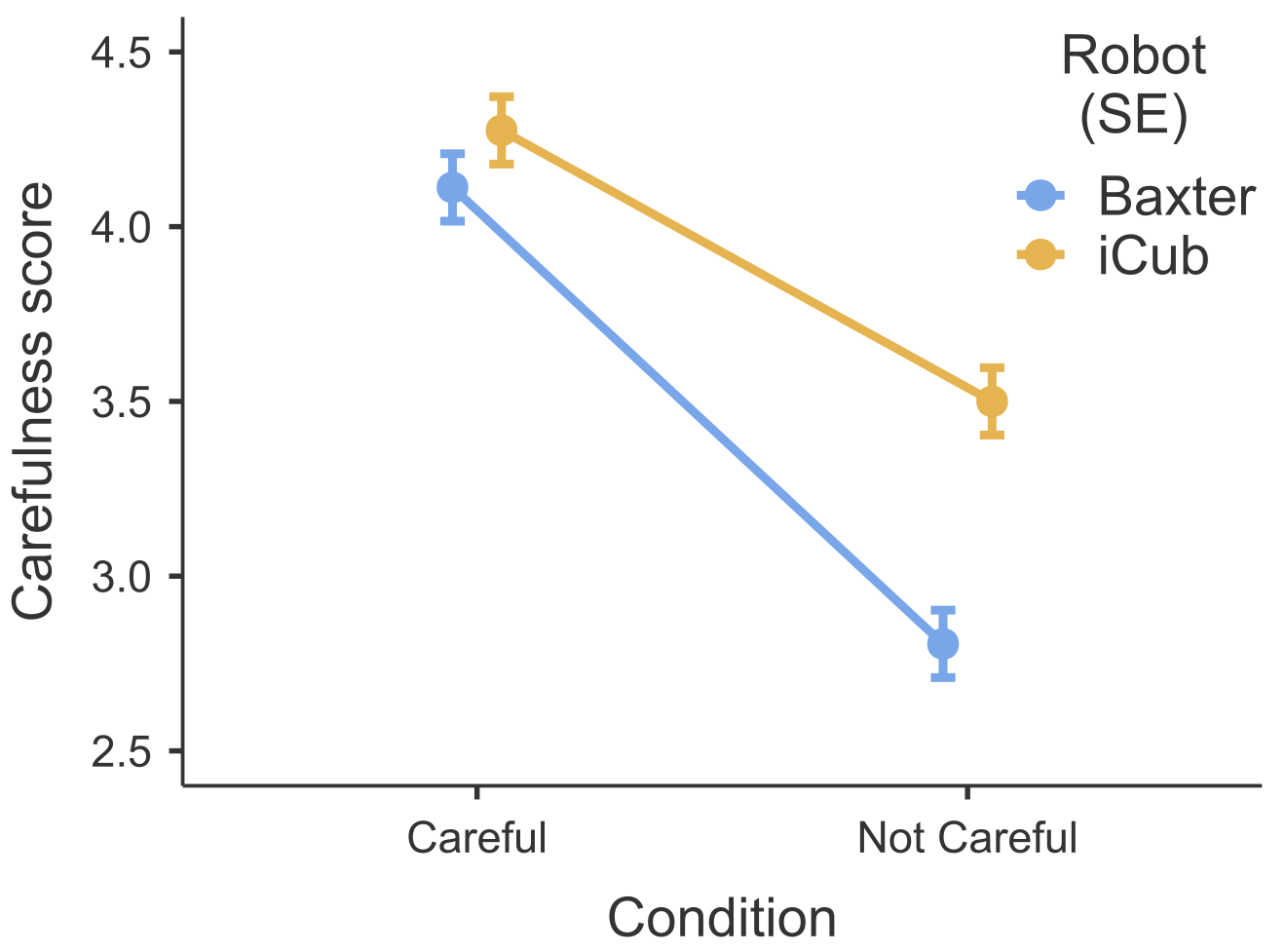}}
\caption{Carefulness perception of Baxter and iCub videos proposed in the questionnaires. Both robots were perceived as significantly more careful when executing the movements planned as careful. Moreover, there is a significant difference in the perception of the two robots in both the conditions, more striking in the Not Careful one}
\label{fig:Cvideos}
\end{figure}

\begin{figure*}[h!]
\centerline{\includegraphics[width=1\textwidth]{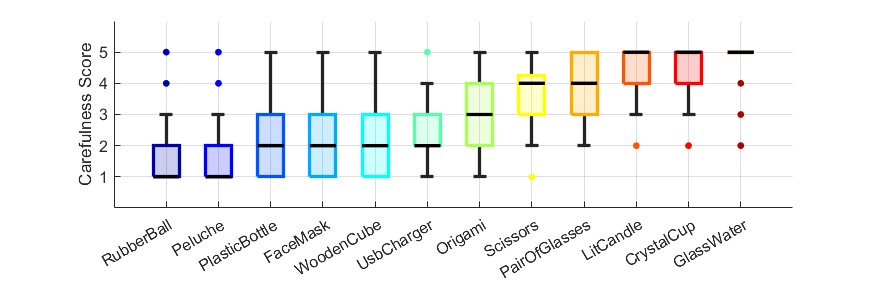}}
\caption{Carefulness score obtained by each item proposed in the questionnaire. The colored boxes represent the 25\textsuperscript{th} and 75\textsuperscript{th} percentiles, the black horizontal line the median. The single dots are outliers}
\label{fig:objects}
\end{figure*}

To assess if the implicit information embedded in the robots movements is communicative of their carefulness, we first evaluated the responses regarding the perceived carefulness in the videos observed by the participants. Considering that two videos were presented for each robot in the C and NC condition, each participant evaluated 8 videos, resulting in a total of 392 carefulness scores, ranging from 1 (NC) to 5 (C). From these data, we ran a mixed model assuming the Carefulness Score as dependent variable, the robot type, the condition and their interaction as factors, and the subjects as cluster variables. The resulting metrics are shown in Figure \ref{fig:Cvideos}. The effect of condition resulted significant ($NC - C$, $estimate = -1.041$, $SE=0.0878$, $t=-11.85$, $p < 0.001$) as well as the effect of the robot ($iCub - Baxter$, $estimate =0.429$, $SE =0.0878$, $t= 4.88$, $p < 0.001$). Moreover, also the interaction between robot and condition is significant ($Careful\cdot Baxter$, $estimate= 0.531$, $SE= 0.1756$, $t= 3.02$, $p = 0.003$). These results indicated that the videos showing what we planned as Careful movements were significantly perceived as more careful, getting a score higher by 1.041. Also, iCub was generally perceived as more careful than Baxter. The carefulness score attributed to Baxter videos covered a wider range, with a stronger difference between the C and NC condition (see Figure \ref{fig:Cvideos}). For both the robots, the Not Careful movements were perceived as rather neutral on the Likert scale.\\
Regarding the carefulness score associated with the proposed 12 items, the results are shown in Figure \ref{fig:objects}. The answers allowed to create a scale going from the objects which were considered not critical to be transported, such as the rubber ball, to the one which requires the highest attention in the manipulation, that is the glass of water full till the brim. 
Bringing together the results of the carefulness scores attributed to the robots' videos, in Figure \ref{fig:Cvideos}, and to the proposed objects, in Figure \ref{fig:objects}, we could evaluate for which items on the list the robots' movements might be considered suitable. On average, none of the robots' movements seem adequate to move the items evaluated as delicate as the \textit{Lit Candle}, onwards. iCub NC movements obtained a Carefulness Score which could match the \textit{Scissors} and \textit{Pair of Glasses}. Even though we did not ask specifically about the item, we can posit that iCub NC movements are suitable, in general, to manipulate objects with an equivalent fragility. Following the same reasoning, Baxter NC attitude seem suitable for those objects which obtained a neutral evaluation, as the \textit{Origami}.
\subsection{In-presence experiment} \label{sec:result_exp}
\begin{figure}[h!]
    \centering
    \includegraphics[width=1\columnwidth]{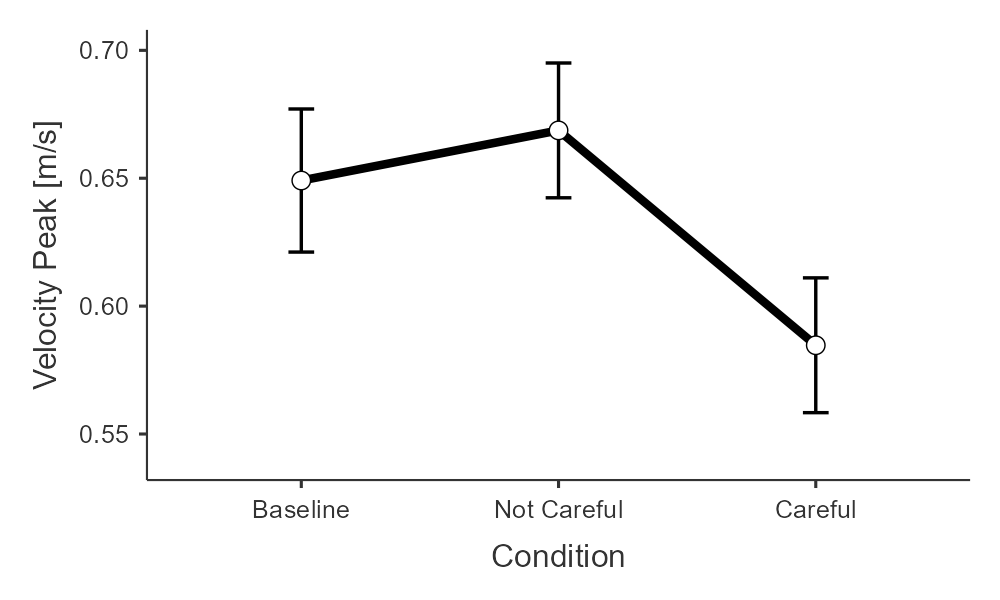}
\caption{Comparison of the maximum velocity reached during the cube transportation in the different phases of the experiment: the baseline, when no stimuli was provided, and the main part where either Careful or Not Careful videos were presented. There is a significant difference between the baseline and the careful condition}
\label{fig:compareBas}
\end{figure}

From the results presented in the previous part, we verified that the movements produced with iCub and Baxter were indeed perceived as more or less careful, coherently with what we planned. The aim of the in-presence experiment was to go a step further and to investigate whether the carefulness modulation of robot motion was sufficient to affect how participants performed their own movements. To answer this question, we computed the hand velocity norm from the components of the velocity, obtained by deriving the three-dimensional position of the considered marker. As metric to quantify the modification of the human movement we considered the maximum velocity reached and the duration of the transport movement performed after watching the video. Indeed, previous studies showed that the velocity peak is a significant measure to distinguish between careful and not careful movements \cite{carefulnessBillard, ICDL2021_GAN}: in particular, careful movements are characterized by lower velocities and longer durations.
\begin{figure}[h]
\centerline{\includegraphics[width=1\columnwidth]{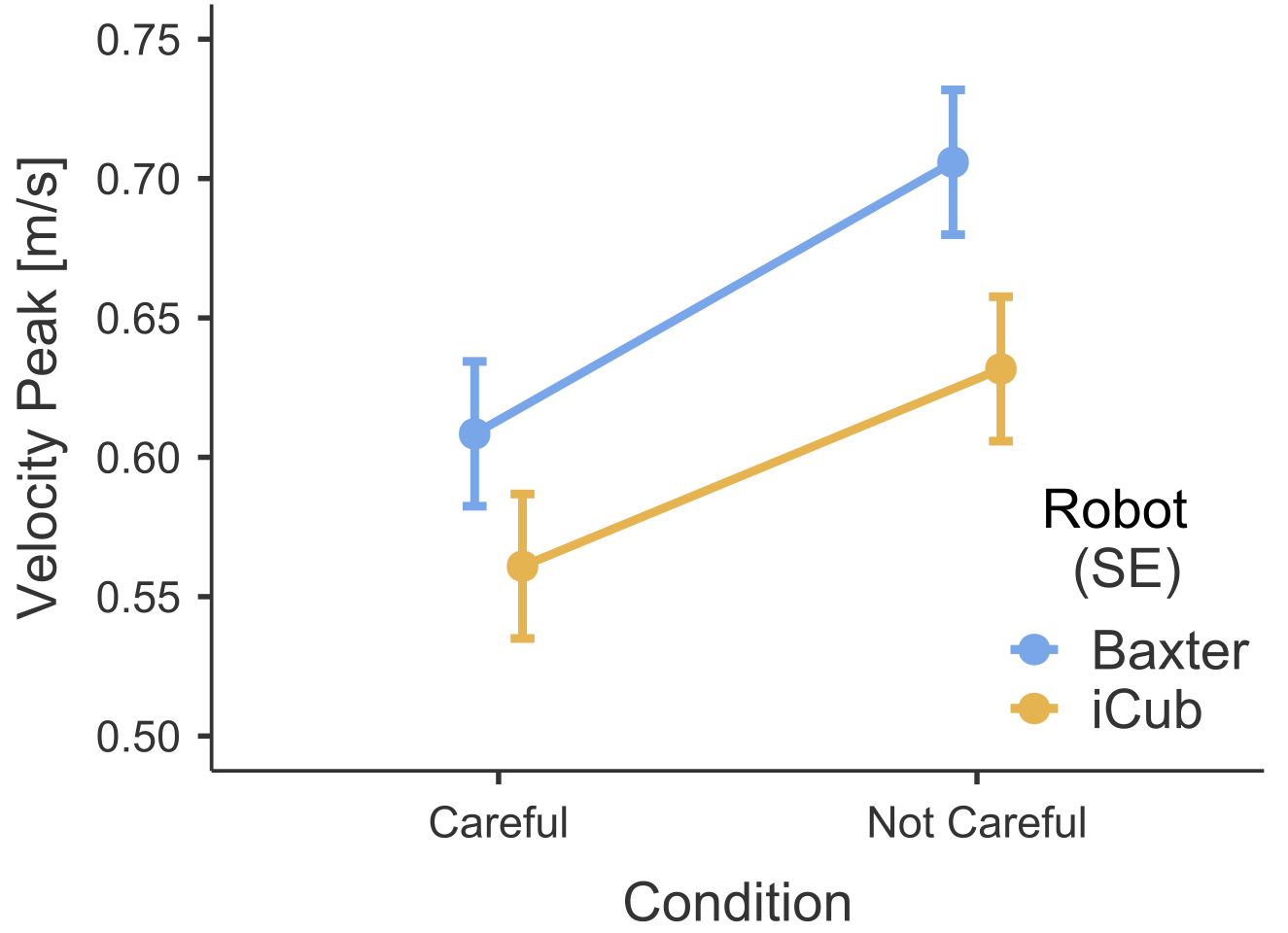}}
\caption{Kinematics modulation of the cube transport movements performed by participants. The hand velocity is significantly modulated by the condition of the video associated to the movement: when the robot performed a Not Careful movement, participants reached a significantly higher maximum velocity than when the robots showed a careful attitude. There is also a significant effect of the robot}
\label{fig:maxvel}
\end{figure}
In the in-presence experiment, six videos were presented for each robot in the C and NC condition, for a total of 24 trials for each of the 11 participants. We ran a mixed model assuming the magnitude of the velocity peak as dependent variable and the robot type, the condition and their interaction as factors, and the subjects as cluster variable. The effect of condition resulted significant ($NC - C$, $estimate = 0.0841$, $SE=0.0127$, $t=6.60$, $p < 0.001$) as well as the effect of the robot ($iCub - Baxter$, $estimate =-0.0608$, $SE =0.0128$, $t= -4.77$, $p < 0.001$). The interaction between robot and condition was instead not significant ($Careful\cdot Baxter$, $estimate= -0.0266$, $SE= 0.0255$, $t= -1.04$, $p = 0.297$). The results are shown in Figure \ref{fig:maxvel}. It can be noticed that the transport movements performed after a careful video stimuli have indeed lower peaks of velocity, with an estimate of the velocity reduction equal to $0.841$ $cm/s$. There is also an effect of the robot, with the videos of iCub soliciting slower movements, with an average reduction in the peak speed of $-0.608$ $cm/s$, respect to Baxter. This is in agreement with what we found from the questionnaires, where iCub movements got a higher carefulness score (see Figure \ref{fig:Cvideos}). For the duration of the movement, we ran the same mixed model using the temporal duration as dependent variable. Again, we found a significant effect of the condition ($NC - C$, $estimate = -0.1796$, $SE=0.0233$, $t=-7.70$, $p < 0.001$) and of the robot ($iCub - Baxter$, $estimate = 0.0886$, $SE =0.0233$, $t= 3.80$, $p < 0.001$), not of their interaction ($Careful\cdot Baxter$, $estimate =0.0758$, $SE =0.0466$, $t= 1.63$, $p = 0.105$). When moving the cube from the starting position to the final one, the covered distance was always of about $30$ $cm$, so the time difference detected depends on how the participants modulated their movement speed. Confirming our hypothesis, after observing careful robot manipulations, participants moved the cube in a slower way, with an estimate increase of duration of $179.6$ $ms$. Analogous considerations were verified as well for other kinematics metrics, such as the mean velocity of the transport or its maximum acceleration. \\
As an additional control, we also compared the movement duration and the velocity peak of the transport after the video stimuli with those of the cube manipulation in the initial baseline, when no videos were presented and participants performed a simple pick and place of the cube. We ran two mixed models, one for each kinematic metrics as dependent variable, the condition (baseline, C or NC) as factor, and the subjects as cluster variable. The results for the velocity peak metrics are shown in Figure \ref{fig:compareBas}. For both the metrics, we obtained a significant difference between the Baseline and the Careful condition (Movement duration, $C-Baseline$: $estimate = 0.1296$, $SE=0.0282$, $t=4.59$, $p < 0.001$; Velocity peak, $C-Baseline$: $estimate = -0.0644$, $SE=0.0163$, $t=-3.94$, $p < 0.001$). This shows that when the robots displayed an attentive attitude, the participants moved significantly slower than they would have naturally. Instead, the models did not evidence a significant difference between the Baseline and the NC condition, even though, as shown in Figure \ref{fig:compareBas}, participants produced slightly faster and shorter movements than during the baseline (Movement duration, $NC-Baseline$: $estimate = -0.0497$, $SE=0.0282$, $t=-1.76$, $p = 0.08$; Velocity peak, $NC-Baseline$: $estimate = 0.0196$, $SE=0.0163$, $t=1.20$, $p = 0.231$).
\section{Discussion}\label{sec:disc}
In the proposed work, we aimed to answer three main research questions: \textit{(RQ1)} assess if the implicit information embedded in the robots' movements is perceived;  \textit{(RQ2)} evaluate the potential modulation produced by the robot movements on how humans handle objects and, \textit{(RQ3)}, investigate whether the robots' embodiment affects how their actions are perceived.\\
Even though the idea of carefulness associated with object manipulation has already been explored, it is difficult to give an unequivocal definition of the term. For instance, unlike the weight, carefulness cannot be measured with a defined scale. From the questionnaire answers in Figure \ref{fig:objects}, about grading the carefulness required for manipulating common-use objects, what emerges is that the concept of carefulness does exist and it is shared among people. Moreover, the emerging gradation correlates with distinct attributes of the items, which may contribute to making an object delicate to handle: the content (glass with water), the material (crystal), the preservation of the object state (lit candle), the value (pair of glasses) or the potential danger (scissors).
The questionnaires outcomes positively confirmed that the movements we produced, with both the robotic platforms, were indeed communicative of the desired feature, that is the presence or absence of a careful attitude in the action (see Figure \ref{fig:Cvideos}). Interestingly, we found an effect of the robot: iCub was perceived as generally more careful in its movements, while Baxter presented a more marked difference between the two conditions. The same result also emerged when comparing the kinematics metrics of the in-presence experiment, presented in Section \ref{sec:result_exp}: iCub videos solicited slower actions by the participants, i.e., more careful manipulation of the cube they found on the table. We can ascribe this to two main factors: how the movements were generated and the embodiment of the robots. As explained in Section \ref{sec:movementGen} and in \cite{specialIssue} with more detail, when controlling the robot end-effector to follow the desired trajectory with a velocity profile of choice, in our implementation, the duration of the original movement is preserved; this means that for longer trajectories the velocity of the movement needs to be higher, whereas for shorter paths the original velocity profile is stretched. The different sizes of the two robots cause the trajectories covered by iCub to be shorter and lower the speed of its movements. iCub's appearance is more human than Baxter's, not only for the hand shape, but also for the kinematic configuration of its arm; this could lead people to identify themselves more with its movements and perceive them, regardless, as more delicate. However, the difference between the C and NC conditions in both robots was significant for the carefulness score attributed to the videos and for the kinematic measures of motor contagion. Hence, even with a platform not designed to have an anthropomorphic shape - such as Baxter - a bioinspired modulation of movement speed effectively communicates motion carefulness.
Finally, Figure \ref{fig:compareBas}, allows to understand better how a C or NC movement is modulated with respect to standard baseline actions. In particular, moving objects in a not careful manner seems to be the natural way of transporting them, whereas we change the movement kinematics when we want to show caution. This seems reasonable to optimize the actions' efficiency and should also be considered when programming robots' behavior.

\section{Conclusion}
This study showed that the movements we generated, associated with object manipulation, are indeed perceived as characterized by different degrees of carefulness for both the robots iCub and Baxter. In particular, the movements we implemented were rated as going from a rather careful attitude to a neutral one. Indeed, what we labeled as Not Careful movements, were perceived as neutral. This comes not by surprise, as confirmed by the comparison with the baseline movements since in everyday life, we tend to act as not careful when manipulating objects, and the kinematics modulation emerges when we are challenged with a difficult task. Moreover, the robots' gestures proved to be efficient in soliciting a motor change in the participants: the simple request to imagine a collaboration with the robot, such as virtually carrying the same object, was enough to bring out a motor adaptation, even along a different direction with respect to the plane where the robot action took place. In a real one-to-one human-robot collaboration, the modulation of human movements in response to the robot ones could become even stronger. \\
Future studies should assess whether the perception of iCub as more careful than Baxter depends on their embodiment or on the different velocities their end-effectors reached in the transport movements. Finally, from the list of items provided in the questionnaires, we plan to evaluate in a more granular manner the features - and their combination - that allows an item to be classified as careful or not to be handled. This to arrive at an explicit definition of the concept of carefulness, aware that an implicit and shared understanding of it already exists. 



\bibliographystyle{IEEEtran} 
\bibliography{references}

\end{document}